\documentclass[11pt]{article}
\usepackage{geometry}
\usepackage{coling2020}
\usepackage{times}
\usepackage{url}
\usepackage{latexsym}
\usepackage{microtype}
\usepackage{graphicx}
\usepackage{url}
\usepackage{float}
\usepackage{makecell}
\usepackage[utf8]{inputenc}
\usepackage[arabic, english]{babel}
\usepackage{adjustbox}

\colingfinalcopy 

\title{A Benchmark Arabic Dataset for Commonsense Explanation}

\author{Saja AL-Tawalbeh, Mohammad AL-Smadi\\Computer Science Department\\ 
Jordan University of Science and Technology\\ 
P.O.Box: 3030 Irbid 22110, Jordan. \\
{\tt sajatawalbeh91@gmail.com, masmadi@just.edu.jo}}

\providecommand{\keywords}[1]{\textbf{\textit{Index terms---}} #1}
\date{}
\begin{document}
\maketitle
\begin{abstract}
Language comprehension and commonsense knowledge validation by machines are challenging tasks that are still under researched and evaluated for Arabic text. In this paper, we present a benchmark Arabic dataset for commonsense explanation. The dataset consists of Arabic sentences that does not make sense along with three choices to select among them the one that explains why the sentence is false. Furthermore, this paper presents baseline results to assist and encourage the future evaluation of research in this field. The dataset is distributed under the Creative Commons CC-BY-SA 4.0 license and can be found on GitHub\footnote{Arabic dataset for commonsense explanation, \url{https://github.com/msmadi/}}.


\end{abstract}

\section{Introduction}
\label{intro}
Introducing the Commonsense to natural language understanding considered as a qualitative change to emphasize the importance of how the machine able to resolve the Commonsense problems. Various datasets have been provided aims to support research in the field of Commonsense knowledge, such as, Choice of Plausible Alternatives (COPA) \cite{roemmele2011choice} presented question dataset related to events and consequences, aims to determine the suitable reason of the premise. Moreover, JHU Ordinal Commonsense Inference (JOCI) \cite{zhang2017ordinal} dataset presented data of five labels beginning from 5 (very likely) to 1 (impossible), which head to determine the reason for humans response after a particular situation. Similarity, Situations with Adversarial Generations (SWAG) \cite{zellers2018swag} which is considered as a large-scale adversarial dataset presented for a sensible commonsense inference that used to determine the expected response after certain action for a particular situation. \cite{ostermann2018mcscript} presented a dataset for a narrative text, that aims to study various questions with their two candidate answers.\\

The recent datasets in the field of commonsense knowledge have been provided by \cite{tawalbeh2020sentence} proposed Arabic dataset for commonsense validation. Moreover, SemEval-2020 \cite{wang2019does,wang2020semeval} has been collected by external reading materials have used to increase the inspiration, for instance, Open Mind Common Sense \cite{havasi2010open} project and take advantage of existing commonsense reasoning questions, for instance, The Winograd schema challenge \cite{levesque2012winograd}, Choice of plausible alternatives \cite{roemmele2011choice}, and SQUABU \cite{davis2016write}. On the other hand, Question Answering (QA) dataset assist the factual commonsense knowledge, for instance, SQUABU \cite{davis2016write} presented scientific questions dataset. Moreover, SemEval-2018 \cite{ostermann2018semeval} presented dataset for machine comprehension using questions commonsense knowledge to determine the correct answer. However, COSMOS QA \cite{huang2019cosmos} presented multiple-choice questions dataset. DCN \cite{Xu2019ABB} proposed deep comatch network for multiple-choice reading comprehension based on BERT pre-trained model. The evaluation done by RACE dataset with significant enhancement. \cite{si2019does} presented a study to investigate the information that BERT can learn from MCRC datasets which are multiple-choice reading comprehension. \cite{saeedi2020cs} used the recent dataset by SemEval to evaluate the presented language model RoBERTa to solve Commonsense validation and explanation task, their work have been conducted by express classification task to multiple-choice task aims to increase the system performance.\\

The lack of the available resources in Arabic domain has encouraged the researchers to assist this domain by providing Arabic datasets. To the best of our knowledge, there is no publicly available dataset regarding commonsense research field. This paper fosters the domain of Arabic Commonsense Explanation (CSE) and provides a benchmark Arabic dataset as well. CSE consists of sentences against the common facts and there true explanation. Furthermore, this paper presented baseline results using the language model methods (i.e., BERT, USE, and ULMFit) to assist and encourage the future evaluation of research in this field. The major idea of the proposed dataset is providing a wrong natural sentence against the real fact with three explanation sentences, however, the system tends to determine the best explanation.
\begin{table}[!h]
\begin{center}
\begin{tabular}{| *{7}{c|} }
    \hline
    File & \multicolumn{2}{c|}{Train} & \multicolumn{2}{c|}{Validation} & \multicolumn{2}{c|}{Test} \\ \hline
    Total & \multicolumn{2}{c|}{10000} & \multicolumn{2}{c|}{1000} & \multicolumn{2}{c|}{1000} \\ \hline

\end{tabular}
\caption{\label{distrbution}The Dataset distribution}
\label{results}
    \end{center}
\end{table}

\begin{table*}[h!]
\centering
\begin{adjustbox}{max width=\textwidth}
\begin{tabular}{|m{0.04\linewidth}|p{0.20\linewidth}|p{0.20\linewidth}|p{0.20\linewidth}|p{0.20\linewidth}|p{0.04\linewidth}|}\hline
    ID & FalseSent & OptionA & OptionB & OptionC & label\\
  \hline
   1089 & \textAR{وضعت جدها في المهد} (She put her grandfather in the cot) & \textAR{الأجداد عادة ما يكونون كبار السن بينما يكون الأطفال صغارًا جدًا} (Grandfathers are usually very old while babies are usually very young) & \textAR{لا يستطيع الجد أن يلد طفلاً} (A grandfather cannot give birth to a baby) & \textAR{البالغ أكبر بكثير من المهد} (An adult is much bigger than a cot) & C \\ \hline
   
    340 & \textAR{دب كان يقود سيارة في كندا} (A bear was driving a car in Canada) & \textAR{الدببة عادة ما تكون بنية اللون} (bears are usually brown) & \textAR{الدب ليس لديه رخصة سيارة وغير قادر على القيادة} (Bear don't have car license and are unable to drive) & \textAR{يمكن لبعض الناس رؤية الدببة تقود سيارة في كندا} (Some people can see bears driving a car in Canada) & B\\ \hline
    
    1785 & \textAR{وضع يخت في السرير} (He put a yacht in bed) & \textAR{اليخت أكبر بكثير من السرير} (A yacht is much larger than a bed) & \textAR{اليخوت عادة ما تكون بيضاء بينما الأسرة بيج} (Yachts are usually white while the beds are beige) & \textAR{هناك سرير على اليخت} (there is a bed on the yacht) & A \\ \hline
    
   
\end{tabular}
\end{adjustbox}
\caption{\label{dataset_EXAMPLE} Selected examples from the provided dataset}
\end{table*}
\section{Dataset Collection and Translation}
\label{data}
The commonsense explanation problem has natural sentence against the fact with three sentences one of them explains why that sentence does not make sense. Each sentence has labeled with the suitable explanation. The original dataset for English language provides by SemEval-2020 \cite{wang2020semeval} \footnote{ \url{https://github.com/wangcunxiang/SemEval2020-Task4-Commonsense-Validation-and-Explanation}} Commonsense Validation and Explanation (ComVE) task which inspired by \cite{wang2019does}. This paper focused on the explanation task for Arabic dataset. Each example in the provided dataset is composed of four sentences: \{s1, o1, o2, o3\} . S1 is a natural sentence against the fact and does not make sense, where o1, o2, and o3 are three options that explain why that sentence does not make sense and the task is to select the most correct out of them.\\

To the best of our knowledge, there is no  publicly available dataset to be used in the research of the commonsense explanation for Arabic language. Based on that, we are providing a benchmark Arabic dataset for the Commonsense explanation problem (why a statement against the natural fact does not make sense). The dataset is provided with 12k rows divided as following: train with 10k , validation with 1k, and test file with 1k (see Table \ref{distrbution}). Each file consists of four columns, the first column contains the false sentence, whereas, the other three columns contain three options to explain why that sentence is wrong. Table \ref{dataset_EXAMPLE} provides examples of the presented Arabic dataset for the commonsense explanation task.  

\section{Experimentation \& Results}
As we aim to present an Arabic benchmark dataset for commonsense explanation, the dataset is provided with a baseline evaluation to address the research task and problem discussed in section \ref{data}. The baseline evaluation is based on several state-of-art transfer based language models i.e. BERT \cite{devlin2018bert}, USE \cite{cer2018universal}, and ULMFit \cite{howard2018universal}. For each false sentence in the test file, the baseline selects the explanation sentence among the three choices provided for this purpose. In order to evaluate baseline models, the accuracy of the approach is measured. We also provided an evaluation code which can be downloaded with the dataset. Future researchers can use the baseline results to evaluate the performance of their proposed research.Table \ref{results} presents the evaluation results for the baseline models.\\  

\begin{table}[h!]
\begin{center}
\begin{tabular}{|c|c|}
\hline \bf Model & \bf Accuracy \\ \hline
Random & 32.5 \% \\
USE  & 33 \% \\
ULMFiT & 32.8 \% \\ 
BERT  & 34.20 \% \\ \hline
\end{tabular}
\end{center}
\caption{\label{results} The experimentation results for the trained baseline models}
\end{table}

As shown in Table \ref{results}, BERT and USE have achieved the highest scores compared to the random results and ULMFiT methods. The results go inline with related work as BERT achieved 45.6\% accuracy for the original English dataset \cite{wang2019does}.


\section{Conclusion}
This paper presents a benchmark Arabic dataset for commonsense explanation. The dataset is distributed under the Creative Commons CC-BY-SA 4.0 license and can be found on GitHub\footnote{Arabic dataset for commonsense explanation, \url{https://github.com/msmadi/}}. We also trained a state-of-the-art transformer-based language models (i.e., BERT, USE, and ULMFit) as baseline research using the benchmark datset. Evaluation results show how challenging is the task the where the best performing baseline model (i.e. BERT) achieved only 34.20\% of accuracy compared to 32.5\% of accuracy for the random model.

\section*{Acknowledgments}
This research is partially funded by Jordan University of Science and Technology, Research Grant Number: 20170107.

\bibliographystyle{coling}
\bibliography{semeval2020}

\end{document}